\documentclass[10pt,twocolumn,letterpaper]{article}

\usepackage{iccv}              % To produce the CAMERA-READY version
\newcommand{\TODO}[1]{\textbf{\color{red}[TODO: #1]}}
\newcommand{\pmh}[1]{\textcolor{violet}{[\textit{#1}]}}
\renewcommand{\TODO}[1]{}

\renewcommand{\pmh}[1]{}

\usepackage{multirow}
\usepackage{pifont}
\usepackage{diagbox}
\usepackage{subcaption}
\usepackage{makecell}

\renewrobustcmd{\bfseries}{\fontseries{b}\selectfont}
\renewrobustcmd{\boldmath}{}
\newrobustcmd{\B}{\bfseries}

\definecolor{iccvblue}{rgb}{0.21,0.49,0.74}
\usepackage[pagebackref,breaklinks,colorlinks,allcolors=iccvblue]{hyperref}

\def\paperID{13870} % *** Enter the Paper ID here
\def\confName{ICCV}
\def\confYear{2025}

\title{Divide and Conquer Self-Supervised Learning for High-Content Imaging}

\author{Lucas Farndale$^{1,2}$, Paul Henderson$^{2}$, Edward W Roberts$^{1,2,*}$, Ke Yuan$^{1,2,*}$ \\
$^{1}$Cancer Research UK Scotland Institute, Glasgow, Scotland\\
$^{2}$University of Glasgow, Glasgow, Scotland\\
{\tt\small \{lucas.farndale, paul.henderson, ed.roberts, ke.yuan\}@glasgow.ac.uk}
}

\begin{document}
\maketitle
\let\thefootnote\relax\footnotetext{\textsuperscript{*}These authors jointly supervised this work}
\begin{abstract}
Self-supervised representation learning methods often fail to learn subtle or complex features, which can be dominated by simpler patterns which are much easier to learn. This limitation is particularly problematic in applications to science and engineering, as complex features can be critical for discovery and analysis. To address this, we introduce Split Component Embedding Registration (\textbf{SpliCER}), a novel architecture which splits the image into sections and distils information from each section to guide the model to learn more subtle and complex features without compromising on simpler features. SpliCER is compatible with any self-supervised loss function and can be integrated into existing methods without modification. The primary contributions of this work are as follows: i) we demonstrate that existing self-supervised methods can learn shortcut solutions when simple and complex features are both present; ii) we introduce a novel self-supervised training method, SpliCER, to overcome the limitations of existing methods, and achieve significant downstream performance improvements; iii) we demonstrate the effectiveness of SpliCER in cutting-edge medical and geospatial imaging settings. SpliCER offers a powerful new tool for representation learning, enabling models to uncover complex features which could be overlooked by other methods.
\end{abstract}    
\begin{figure}
    \centering
    \includegraphics[width=\linewidth]{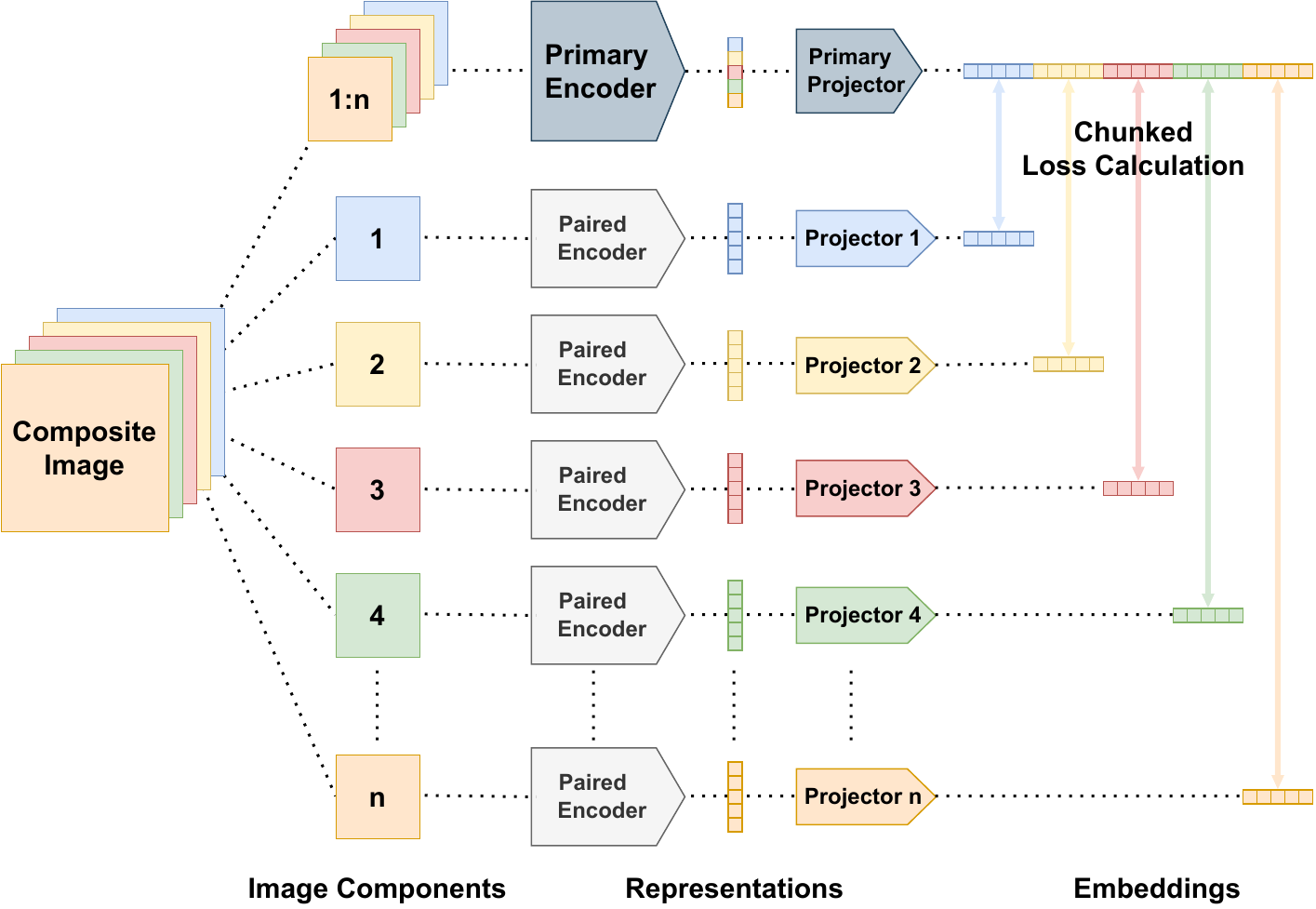}
    \caption{SpliCER training architecture}
    \label{fig:split-rep-architecture}
\end{figure}

\section{Introduction}
\label{sec:intro}

Recent advances in our ability to acquire highly detailed, information-rich imaging at scale have resulted in high-content imaging -- imaging which aims to maximise data capture -- being widely used in science \cite{way2023evolution}, medicine \cite{radtke2022ibex,lin2023high}, and engineering \cite{xia2010structural}. In general, however, computer vision methods are designed for natural images such as ImageNet \cite{imagenet_cvpr09}, which can be well-described with relatively simple features \cite{singla2021salient}. High-content imaging, such as advanced medical imaging or satellite imagery, generally contains subtle or complex features which can be difficult to distinguish from noise without strong supervision. However, subtle or complex features can be among the most interesting features for scientific discovery, such as the interaction or morphology of subtly different cell types \cite{hale2024cellular}, as simpler features are generally more readily discovered by humans. It is therefore essential that alternative methods for learning these features are explored and developed. A prime example is multiplex imaging. Unlike typical RGB images, these images contain more than three channels, each with a distinct meaning, creating significant variation in the complexity of the features in each channel. Despite their prevalence, multiplex images are generally understudied in computer vision, and largely lack dedicated methods to extract key information stored in them. The effect of this can be that some channels dominate the image, and models ignore features from subtler channels.

In many cases, we can leverage an intuitive reason or some prior knowledge indicating that the features of a region or channel of an image are of particular interest, or could be ignored. It is usually possible to split the image into smaller components, perhaps by generating a segmentation mask or cropping the image, and this information could be used to guide the model to learn features from each component \cite{farndale2023more,farndale2023trident,farndale2024synthetic,nakhli2024volta}. Ensuring features are distributed across all components could ensure that complex features which cluster in a certain component will be learned. However, we generally lack the methods to enable this type of \emph{divide and conquer} strategy for high-content imaging.

In this work we introduce a novel representation learning architecture, \emph{SpliCER}, which deconstructs an image into components, and distils information from each component to direct models to learn features from each part, significantly improving representation learning and downstream performance. This forces the model to learn features from each component, meaning components with more complex or subtler features cannot be ignored. As this deconstruction is only performed during training, the encoder can then be used downstream without requiring deconstruction. SpliCER is able to utilise any self-supervised architecture or loss, so it is generally applicable across self-supervised methods.

\begin{figure*}
    \centering
    \includegraphics[width=0.8\linewidth]{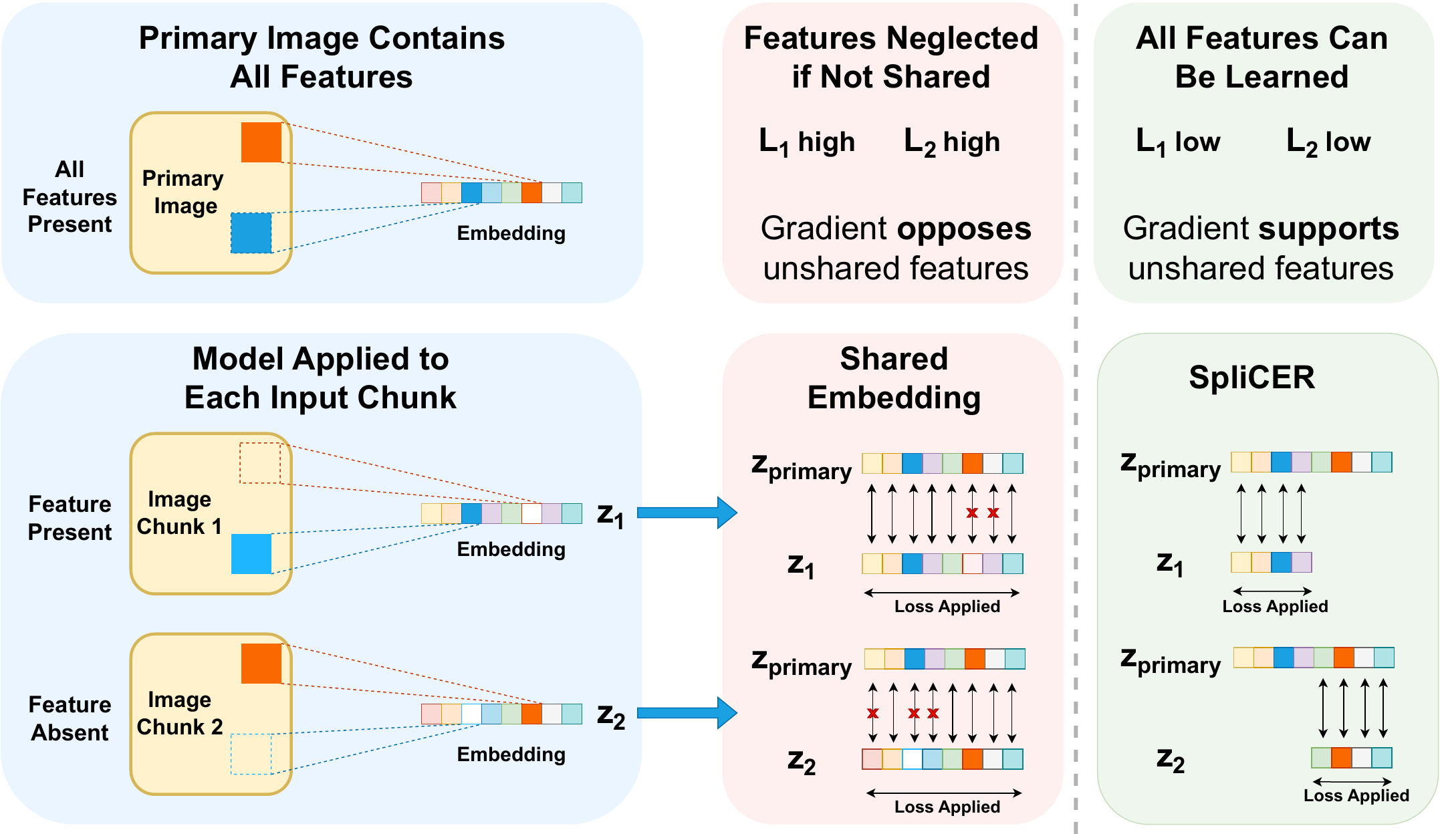}
    \caption{Schematic of the benefits of SpliCER compared to mapping multiple inputs to a shared latent space.}
    \label{fig:shared-latent-space}
\end{figure*}
\section{Background and Related Work}
\label{sec:background}

Self-supervised learning (SSL) uses inherent structure in data to learn useful representations without relying on manual labels. This is achieved using a proxy objective which directs the model to learn features satisfying certain conditions \cite{shwartz2023compress}, without prior knowledge of the desired downstream task(s). Recent work demonstrates that many of the most effective methods map multiple views of the same input into a \emph{joint embedding} (JE) \cite{giakoumoglou2024review}, maximising the mutual information shared between views. Typically, this is achieved through augmenting the input to artificially create different views, with augmentations designed to avoid augmenting the key semantic features which the model should learn \cite{bachman2019learning}.

\subsection{Simplicity Bias and Domain Generalisation}

Representation learning relies on proxy metrics to determine which features should be learned. Typical methods seek to optimise desirable properties of the representation, such as maximising the mutual information between embeddings \cite{chen2020simple,ozsoy2022self}, or minimising the covariance between features \cite{bardesvicreg}. \emph{Simplicity bias} \cite{shah2020pitfalls} is the tendency of machine learning models to learn the simplest features required to satisfy any given objective \cite{robinson2021can,chen2021intriguing}, deteriorating the quality of models' learned representations. Models must seek to reduce redundancy in their representations so learning every possible complex feature is not possible, as these could be mostly noise. However, in many cases, learning the simplest features is a \emph{shortcut solution} that neglects the complexity of the true underlying manifold \cite{geirhos2020shortcut}.

The encoder architecture has been shown to have little effect on this bias \cite{pinto2022impartial}. A natural avenue for development is to incorporate additional sources of information into training which make it easier to identify these features. Whilst not explicitly designed to solve simplicity bias, recent work has established the effectiveness of this method, such as vision-language modelling \cite{radford2021learning} and knowledge distillation from paired data \cite{farndale2023trident}. However, these additional sources of information may not always be available.

Shortcut solutions are often exposed with transfer learning, as these features typically do not generalise as well as solutions which capture the complexity of the underlying manifold \cite{vasudevamitigating}. Sometimes simpler features are desirable in this setting, as they are usually more robust to minor changes in appearance. However, many simple features such as texture or background can be artefacts of a training set and consequently are harmful to domain generalisation.

\subsection{Self-Supervised Learning Feature Complexity}

It has been shown that self-supervised models learn features with a variance greater than that imposed by the augmentation regime, as these are the only features which can be reliably predicted under augmentation \cite{jing2021understanding}. In this setting, features which either only exhibit small changes in appearance or have larger, infrequent changes, will typically be very hard to differentiate from noise induced by augmentation. This reduces overfitting on trivial or undesirable features, but also results in models failing to learn the potentially large space of features which are valuable but have low variance. Naturally, features can not be easily normalised to unit variance, as they are not known a priori. For example, these interesting low-variance features could be minor changes in shape which are hidden by elastic deformations induced by the augmentation regime such that the model does not learn to identify them. If there are few features with sufficient variance to be learned relative to the augmentation regime, this phenomenon can lead to dimensional collapse, where a model produces a low-rank representation \cite{jing2021understanding} and deteriorates performance \cite{wang2022contrastive}.

Sridharan and Kakade \cite{sridharan2008information} showed that retaining predictive features while discarding superfluous information requires a \emph{multi-view assumption} about the relationship between the views and downstream labels. This is the assumption that for two SSL inputs $x_1,x_2$, the meaningful semantics which should be learned are shared by both inputs (views) \cite{sridharan2008information}. Formally, we write that for two views $x_1,x_2$ and some task label $y$, there exists some $\epsilon_{\mathrm{info}}>0$ such that
\begin{equation}
\label{eq:multi-view-assumption}
    I(y;x_1|x_2)\leq\epsilon_{\mathrm{info}}, \quad I(y;x_2|x_1)\leq\epsilon_{\mathrm{info}}.
\end{equation}
This is the foundational assumption underpinning self-supervised JE architectures \cite{shwartz2023compress}. Several attempts have been made to relax this assumption \cite{kahana2022contrastive,wang2022rethinking}, although these rely on reconstructing the input image, which results in learning redundant features such as precise landmark locations and colours \cite{balestriero2024learning}, and consequently producing worse representations for downstream tasks in general. Thus, designing an architectural framework for relaxing the multi-view assumption is an open problem \cite{shwartz2023compress}.

\subsection{Minimal Sufficient Representation}

Representation learning is the task of finding a mapping $f$ from an input $x$ to a \emph{representation} $z$ which is informative about some feature(s) of the data $y$. A \emph{sufficient} representation of $x$ for $y$ is one which is as informative about $y$ as the input data itself, with no loss of information due to the encoding. Formally, this requires that the conditional mutual information $I(x;y|z)=0$ \cite{wang2022rethinking}.

\citealt{federici2020learning} show that conditional mutual information can be subdivided into predictive and superfluous information using the chain rule of information theory:
\begin{equation}
\label{eq:chain-rule-sup}
    I(x;z) = \underbrace{I(x;z|y)}_{\text{superfluous information}} + \underbrace{I(y;z)}_{\text{predictive information}}.
\end{equation}
The \emph{minimal sufficient representation} is the sufficient representation that most minimises the first term. In self-supervised learning, the label is assumed to be a priori unknown. Multi-view SSL methods therefore seek to minimise redundancy, with augmentations designed such that any information not shared between views will be an artefact of the augmentation, not helpful for a future predictive task. An input $x$ is augmented in two different way to give augmented inputs $x_1,x_2$, which are then mapped by the encoder $f$ to representations $z_1,z_2$. A projector $g$ then maps $z_1,z_2$ to embeddings $e_1,e_2$, on which the loss is applied. Reformulating \cref{eq:chain-rule-sup} for the unsupervised setting gives
\begin{equation}
    \label{eq:chain-rule-unsup}
    I(x_1;e_1) = \underbrace{I(x_1;e_1|x_2)}_{\text{superfluous information}} + \underbrace{I(x_2;e_1)}_{\text{predictive information}}.
\end{equation}
In practice, it is not necessarily desirable for an encoder to learn a strictly sufficient representation, as this may only be possible by retaining an undesirable amount of superfluous information. As such, solutions typically seek to minimise the relaxed Lagrangian objective $\mathcal{L}=\mathcal{L}_{1\rightarrow 2}+\mathcal{L}_{2\rightarrow 1}$, where
\begin{equation}
\label{eq:relaxed-lagrangian}
    \mathcal{L}_{1\rightarrow 2}(\lambda_{12}) = I(x_1;e_1|x_2)-\lambda_{12} I(x_2;e_1),
\end{equation}
where $\lambda_{12}$ is the Lagrange multiplier induced by the optimisation problem, controlling the trade-off between the amount of superfluous information learned by the model and the amount of predictive information not learned, and $\mathcal{L}_{2\rightarrow 1}$ is defined symmetrically \cite{federici2020learning}.

\subsubsection{TriDeNT and Multiple Latent-Space Models}

\emph{TriDeNT} \cite{farndale2023trident} is a model which was developed to address the restrictions of the multi-view assumption by adding an additional branch to the model architecture to create three simultaneous joint-embeddings. This allows the model to trade off learning simple features against complex features which receive a strong supervisory signal from the paired data. For a joint-embedding loss $\mathcal{L}$, TriDeNT minimises $\sum_{i, j \in \{1,2,*\}}\mathcal{L}(e_i,e_j)$. TriDeNT is effective for distilling information in a medical image setting, with significant performance improvements over baseline methods.

Despite TriDeNT's success with using knowledge distillation to guide the model to relevant, low-variance features, it is limited in scope by only being able to utilise one source of paired data. Additional joint-embeddings can overconstrain the model, leading to worse performance, as features shared between additional inputs but not present in the primary input receive a strong supervisory signal that does not relate to anything seen by the encoder. Alternatively, additional inputs may share few or no features, and cause the encoders to limit the variance of their learned features to avoid large penalties from this mismatch. A natural approach might be to create joint embedding only between the primary branch and each additional branch in order to limit the effect of additional branches on each other, however, it has been shown that this still leads to emergent alignment between branches \cite{girdhar2023imagebind}.

\subsection{Multiplex Imaging}
\label{sec:multiplex-background}

Widely used in domains as varied as medicine \cite{kobayashi2010multiplexed}, biology \cite{lewis2021spatial}, and geoscience \cite{drusch2012sentinel}, \emph{multiplex/multichannel} images are \emph{hyperstacks} of spatially registered greyscale images each containing different information about the same scene or object. Unless colour is important, swapping the channels of an RGB image does not significantly change its semantic content, but swapping the order of channels in a multiplexed image will greatly change their meaning.

\subsubsection{Spatial Proteomics}

\emph{Spatial proteomics} are a range of methods used to analyse the spatial distribution of proteins in tissue samples, winning Nature Methods' Method of the Year 2024 \cite{karimi2024method}. These techniques generate multiplexed images where each channel represents the localisation of a specific protein, revolutionising researchers' capacity to understand the morphology and interactions of different cell types. The proteins studied in different experiments are rarely the same, so each dataset is likely to feature a different number of channels, each with different meanings. Interpretation of these images requires extensive knowledge of each protein and how it is typically distributed. Interpretable composite images which map each channel to a colour (\eg \cref{fig:multiplex-imaging}) quickly become intractable with an increasing number of channels, making human interpretation extremely challenging.

Despite this, the volume of multiplex spatial proteomics data being created is rapidly expanding, meaning that effective computational methods for its analysis are urgently needed. Typical analysis reduces each cell to a vector of mean expression in each channel \cite{brbic2022annotation,shaban2024maps}, neglecting morphology, secreted proteins, and extracellular matrix components \cite{bussi2024multiplexed}. There is a paucity of manually labelled datasets, meaning that supervised approaches have limited efficacy, and can only reproduce known phenotypes. 

\subsubsection{Hyperspectral Imaging}

As humans can only see light in the three bands of the visible spectrum -- red, green and blue -- imaging is typically only performed in this range of frequencies. However, the rest of the electromagnetic spectrum contains rich information that can be utilised for applications from food processing \cite{gowen2007hyperspectral} to urban planning \cite{weber2018hyperspectral} and atmospheric monitoring \cite{stuart2019hyperspectral}. A key application is land-use classification, where hyperspectral bands from satellite images contain information such as cloud cover, vegetation health, geology, and soil moisture content \cite{drusch2012sentinel}. These images are rapidly produced at scale, with the Sentinel-2 satellites covering the entire surface of the earth approximately every five days \cite{drusch2012sentinel}. Machine learning tools which can fully leverage the vast amounts of data generated by these satellites could discover critical information to combat climate change and understand changing patterns of land use \cite{prexl2023multi}. However, like medical imaging, different bands have significantly different feature complexity, meaning critical subtle or complex information may be ignored.

\begin{figure*}[t]
\centering
\begin{subfigure}{0.22\textwidth}
\includegraphics[width=\linewidth]{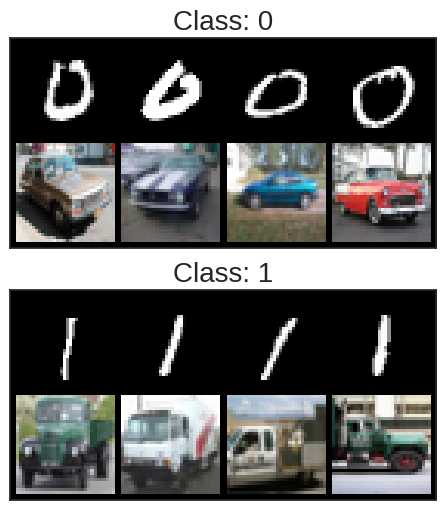}
\caption{}
\label{fig:mnistcifar-examples}
\end{subfigure}
\begin{subfigure}{0.32\textwidth}
\includegraphics[width=\linewidth]{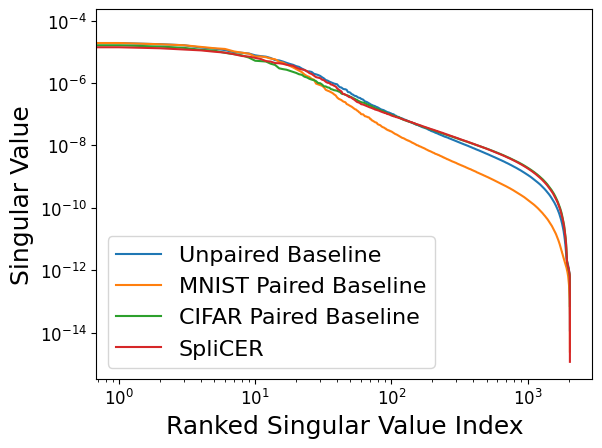}
\caption{}
\label{fig:rank}
\end{subfigure}
\begin{subfigure}{0.45\textwidth}
    \scalebox{0.7}{
    \begin{tabular}{cccccc}
    \toprule
         \multirow{2}{*}{Loss} & \multirow{2}{*}{Paired Image} & \multirow{2}{*}{Method} & rNONE & rMNIST & rCIFAR \\
         & & & (All) & (Complex) & (Simple) \\
    \midrule
    \multirow{10}{*}{VICReg} & \multirow{2}{*}{NONE} & TriDeNT & 0.9954 & 0.6230 & 0.9939 \\
    & & Baseline & 0.9908 & 0.6410 & 0.9918 \\
    \cmidrule{2-6}
    & \multirow{3}{*}{MNIST} & SpliCER & 0.9974 & \B 0.6425 & 0.9974 \\
    & & TriDeNT & \B 0.9985 & 0.4990 & \B 0.9980 \\
    & & Baseline &   0.9980 & 0.5270 & \B 0.9980 \\
    \cmidrule{2-6}
    & \multirow{3}{*}{CIFAR} & SpliCER & \B 0.9888 & \B 0.6840 & \B 0.9852 \\
    & & TriDeNT & 0.8699 & 0.6645 & 0.8041 \\
    & & Baseline & 0.7372 & 0.6830 & 0.7020 \\
    \cmidrule{2-6}
    & \multirow{2}{*}{BOTH} & SpliCER & \B 0.9964 & \B 0.6675 & \B 0.9969 \\
    & & $\Sigma$-JE & 0.9934 & 0.6530 & 0.9934 \\
    \bottomrule
    \end{tabular}
    }    
\caption{}
\label{tab:rMNIST}
\end{subfigure}

\caption{(a) MNIST-CIFAR examples (b) Ranked singular values of models trained on MNIST-CIFAR (c) SSL performance for models trained and evaluated on randomised MNIST (rMNIST), randomised CIFAR (rCIFAR), or with no randomisation (rNONE).}
\label{fig:MNIST-CIFAR}
\end{figure*}

\section{Methodology}

As there are few established methods for our problem setting, in \cref{sec:preliminaries} we first consider some preliminary elements motivating the design of SpliCER and justify why these form strong baselines. We then move on to the technical contributions of SpliCER in \cref{sec:splicer}.

\subsection{Preliminaries}
\label{sec:preliminaries}

The first baseline we use in each experiment is a standard self-supervised architecture, either VICReg \cite{bardesvicreg} or SimCLR \cite{chen2020simple} (denoted \emph{Baseline}). We use ResNet \cite{he2016deep} models for each input, with the number of channels determined by the number of image channels. Full details are provided in \cref{sec:training}. These models receive two augmentations of the same input, and maximise the mutual information between branches. Existing methods for knowledge distillation from paired data in self-supervised learning generally use these same joint-embedding architectures to map inputs to the same latent space \cite{radford2021learning,farndale2023trident,girdhar2023imagebind}. While the multi-view assumption is not especially restrictive for standard, multi-view approaches, these approaches become problematic, because  information not shared between inputs can be ignored, degrading performance \cite{farndale2023trident}. This is because the model loses its implicit supervisory signal for all features which are not present in the paired information. For example, it is impossible to completely describe the semantics of an image, so a vision/language model will have a considerable amount of the content of the image not shared between inputs. Similarly, an image could be equally well-described with poetry or a technical description. Neglecting this unshared information represents a major limitation of methods such as CLIP \cite{radford2021learning}. 

Consequently, information in inputs cannot be divided as neatly into superfluous and predictive information based on being shared between inputs. From \cref{eq:multi-view-assumption} we can see that any information not shared between views is superfluous under mutual-information maximisation, hence additional modalities may stop the model learning useful unshared features. This is seen in ImageBind \cite{girdhar2023imagebind}. Despite the emergence of alignment between modalities with no paired training data, the performance of the model is considerably worse than standard approaches. Nevertheless, aligning multiple inputs by mapping into shared latent spaces remains a promising approach for finding emergent alignment between feature sets, so we include this as a baseline (\textbf{$\Sigma$-JE}), to study the aggregation of different joint embeddings. The input to the primary branch is the original image, while the input to each additional branch is a component of the image. Where applicable we also include TriDeNT, however, there are few scenarios considered where there is only one source of paired data.

\begin{figure*}
\centering
\begin{subfigure}{0.3\textwidth}
    \centering
    \includegraphics[width=\linewidth]{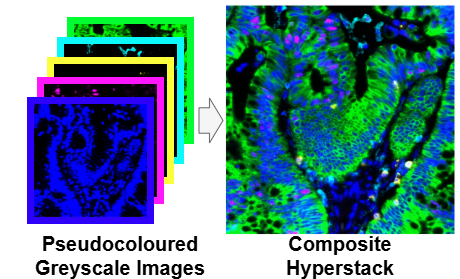}
    \caption{}
    \label{fig:multiplex-imaging}
\end{subfigure}
\begin{subfigure}{0.4\textwidth}
\centering
\scalebox{0.8}{
    \begin{tabular}{cccc}
    \toprule
        Loss & Method & \makecell{Cell Types\\(Simple)} & \makecell{T Cells\\(Complex)} \\
    \midrule
        \multirow{3}{*}{VICReg} & SpliCER & \B 0.9003 & \B 0.8136 \\
        & $\Sigma$-JE & 0.6715 & 0.5222 \\
        & Baseline & 0.8820 & 0.5665 \\
    \midrule
        \multirow{3}{*}{SimCLR} & SpliCER & 0.8950 & \B 0.8469 \\
        & $\Sigma$-JE & 0.8407 & 0.6886 \\
        & Baseline & \B 0.8972 & 0.5539 \\
    \bottomrule
    \end{tabular}
    }
    \caption{}
    \label{tab:multiplex-accuracy}
\end{subfigure}
\begin{subfigure}{0.25\textwidth}
    \centering
    \includegraphics[width=\textwidth]{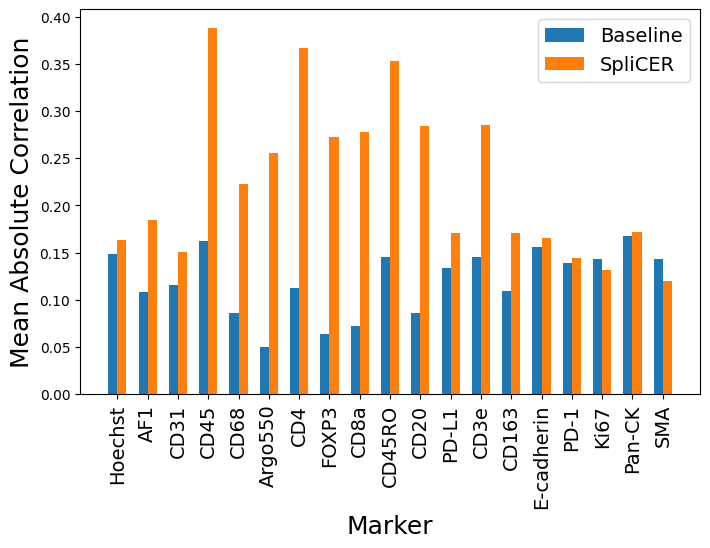}
    \caption{}
    \label{fig:multiplex-xcorr-orion}
\end{subfigure}
% \begin{subfigure}{0.5\textwidth}
%     \centering
%     \includegraphics[width=\textwidth]{figs/xcorr-cellneighs-all.png}
%     \caption{}
%     \label{fig:multiplex-xcorr-cellneighs}
% \end{subfigure}
\caption{(a) Example of a multiplex immunofluorescence hyperstack (b) Orion-CRC classification accuracy (c) Mean absolute cross correlation between representation elements and marker intensities on Orion-CRC for baseline and SpliCER}
\label{fig:multiplex-results}
\end{figure*}

\subsection{SpliCER: Split Component Embedding Registration}
\label{sec:splicer}

Despite the difficulty in creating manually labelled datasets for supervised learning, it can be relatively simple to systematically identify the image component containing a feature of interest by creating segmentation masks or other identifiers using generalist pretrained models such as Segment Anything \cite{kirillov2023segment}. In particular, multiplex images can be easily deconstructed into their constituent channels. Methods which can leverage this information without overconstraining the model by enforcing pairwise embedding alignment will therefore be able to train models which identify low-variance features in images and achieve better downstream performance.

We introduce \emph{Split Component Embedding Registration (SpliCER)}, a self-supervised training architecture which allows models to flexibly learn features from all components of a deconstructed image while avoiding shortcut solutions. SpliCER optimises this process by i) mapping the primary (not deconstructed) image to an embedding, ii) mapping each component of the image to a distinct embedding, iii) splitting the primary image's learned embedding into \emph{chunks}, with the embedding of each component registered to a distinct chunk of the primary embedding.

This creates a separate joint embedding between the embedding of each deconstructed input component and a chunk of the primary image's embedding. \cref{fig:shared-latent-space} illustrates the intuition behind this. When the embeddings of all branches must be aligned to the entire primary embedding, this necessitates that branches which do not contain a feature must align their embedding to branches with that feature. These embeddings therefore contain random noise or collapse to a constant, resulting in gradients which oppose learning that feature in the primary embedding. In contrast, SpliCER aligns only a portion of the primary embedding to each branch's embedding, resulting in gradients not opposing features which are only present in one or few branches.

Consider a primary input $\bar{x}$, with privileged inputs $x_1^*,\ldots,x_n^*$, a learned primary representation $\bar{z}$, and privileged representations $z_1^*,\ldots,z_n^*$, mapped into embeddings $\bar{e}\in\mathbb{R}^N$ and $e_1^*,\ldots,e_n^*\in\mathbb{R}^M$ respectively. Splitting $\bar{e}$ into $n$ chunks such that $\bar{e}=\bar{e}_1||\bar{e}_2||\ldots||\bar{e}_n$, we can construct an optimisation problem
\begin{equation}
\label{eq:split-lagrange}
\begin{split}
    \mathcal{L} = \sum_{i=1,\ldots,n} & \mathcal{L}_{i}(\lambda_{i}) \\
    = \sum_{i=1,\ldots,n} & I(x_i;e_i|\bar{x}) + I(\bar{x};\bar{e}|x_i) \\
    & - \lambda_{i}\left[I(\bar{x};e_i) - I(x_i;\bar{e})\right] \\
\end{split}
\end{equation}
following \cref{eq:relaxed-lagrangian}. We assume that all chunks are of equal size for brevity, but this is not a necessary restriction. Note that different chunks of the embedding are optimised separately, meaning that there is less risk to the model predicting low-variance features from the privileged inputs, as these do not come at the cost of features from a different privileged input, and there is no need for different chunks to be aligned in embedding space. Note also that this formulation removes the reliance that all inputs are jointly optimised, as each branch is only optimised relative to the primary branch. As the primary embedding $\bar{e}$ will, in general, have a significantly greater dimension than the primary representation $\bar{z}$, an optimal model should replicate features across different embedding elements. This means that a large amount of features which are replicated across different stains can be condensed into a few representation elements, and the signal from low-variance stains can force other subtler features to be learned.

\section{Experiments}
\label{sec:methodology}

We provide full descriptions of the datasets and training hyperparameters used in \cref{sec:datasets,sec:training} respectively. We also provide model ablations in \cref{sec:ablations}.

\begin{figure*}[t]
    \centering
    \begin{subfigure}{0.55\textwidth}
        \centering
        \includegraphics[width=\textwidth]{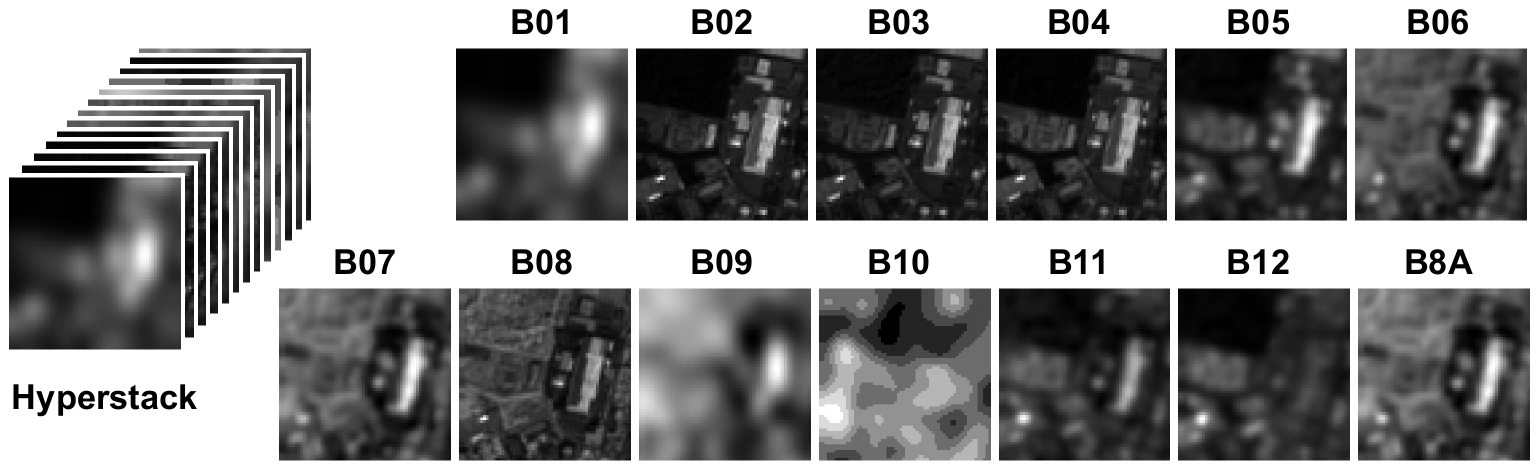}
        \caption{}
        \label{fig:hyperspectral-example}
    \end{subfigure}
    \begin{subfigure}{0.44\textwidth}
        \centering
        \begin{tabular}{ccccc}
        \toprule
        Loss & Method & EuroSAT & MMEarth \\
        \midrule
        \multirow{2}{*}{VICReg} & SpliCER & \B 0.8921 & \B 0.7015 \\
        & Baseline & 0.8481 & 0.4976 \\
        \midrule
        \multirow{2}{*}{SimCLR} & SpliCER & \B 0.9044 & \B 0.7061 \\
        & Baseline & 0.8479 & 0.5986 \\
    \bottomrule
    \end{tabular}
    \caption{}
    \label{tab:eurosat-results}
    \end{subfigure}
    \caption{(a) Example of a hyperspectral image hyperstack (b) Hyperspectral imaging classification accuracy}
    \label{fig:eurosat}
\end{figure*}

\subsection{MNIST-CIFAR Simplicity Bias Evaluation}

Our first experiment compares the complexity of features learned by SpliCER to those learned by baselines. We use the MNIST-CIFAR dataset, an established benchmark for simplicity bias, to assess whether the model learns \emph{simple} features, \emph{complex} features, or both. The dataset is an amalgam of the MNIST handwritten digit dataset \cite{lecun1998gradient}, which contains simple features, and the CIFAR-10 dataset, which contains more complex features. Each image in the MNIST-CIFAR dataset is constructed by concatenating an MNIST image to a CIFAR image with the same label, such that the model concurrently processes both. Classes are mapped such that 0 is always car and 1 is always truck, so a model can find shortcut solutions by only learning MNIST features, as these are just as informative about the label as the CIFAR features, but are easier to learn.

We evaluate which features have been learned by training a classifier while randomising either the MNIST or CIFAR images, denoted as rMNIST and rCIFAR respectively, with rNONE being unrandomised. On rMNIST, a model which has learned a complete shortcut solution will have accuracy around 50\%, as it will not be able to make predictions from the CIFAR features. On rCIFAR, this is unlikely to have an effect on a model which has learned a shortcut solution. Models which have learned more complex features will be less affected by MNIST randomisation and more affected by CIFAR randomisation.

In \cref{tab:rMNIST} we demonstrate that baseline SSL training does not learn a complete shortcut solution, with accuracy of 64\% on rMNIST and near perfect ($\sim$100\%) performance on rNONE/rCIFAR. However, this indicates that predictive features have not been learned, as a model trained on the CIFAR data alone achieves 72.6\% accuracy. \cref{fig:rank} demonstrates the effect of the choice of paired data on the learned features. Pairing MNIST results in a lower rank representation, while the model trained with CIFAR as paired data has similar rank to unpaired training and SpliCER. We pair either the MNIST image, the CIFAR image or both, to compare the efficacy of different distillation methods. We find that SpliCER can make up to a 4 percentage point improvement when given CIFAR as privileged information. This implies that the signal from the CIFAR branch guides the model to learn more complex features. 

In baseline or TriDeNT, using MNIST as privileged information appears to cause the model to focus even more on the simple features and collapse to a shortcut solution, resulting in poorer performance on rMNIST, but marginally increasing performance on rCIFAR. Similarly, using CIFAR as privileged information increases performance on rMNIST, but heavily degrades performance on rCIFAR. This is because the privileged information forces the model to primarily pay attention to the shared features, at the expense of the unshared features.

SpliCER achieves similar performance to the baseline and TriDeNT on the task that corresponds to the privileged information, but also completely mitigates the performance degradation observed in the other models. This implies that SpliCER can benefit from being guided by the paired data, but the chunking of its embedding does not restrict its ability to learn the remaining features. Achieving a classification accuracy of 68.4\%, the SpliCER approaches the performance of a model trained on the CIFAR images alone, which achieves an accuracy of 72.6\%, without the option to learn the simple features in MNIST.

\subsection{Spatial Proteomics}

We next demonstrate the real biomedical applicability of SpliCER to multiplex immunofluorescence images. As there are no established benchmarks for multiplex immunofluorescence, we construct three datasets from the Orion-CRC dataset \cite{lin2023high}: a pretraining dataset subsampled from all cell types with a distribution reflecting normal tissue distribution; Cell Types - a balanced evaluation dataset for classifying different cell types which require relatively simple features; and T Cells - a more fine-grained balanced evaluation dataset for classifying subtypes of T cells, requiring more complex features. As discussed in \cref{sec:multiplex-background}, there are significant differences in the variance of features in different channels of multiplex image, as shown in \cref{fig:multiplex-variance}. The Cell Types task requires models to learn both simple and complex features, while T Cells assesses the whether the model has learned features of the low variance markers CD4, CD8, and FOXP3.

Consistent with the MNIST-CIFAR tasks, we find that SpliCER significantly improves performance for learning the complex features without degrading performance on the simple features. This leads to a consistent performance of around 89\% on the cell type classification task, and a large improvement from 57/55\% to 81/85\% on the T cell subtyping task for VICReg/SimCLR respectively. Furthermore, we show in \cref{fig:multiplex-xcorr-orion} that SpliCER learns features with a considerably greater correlation with the marker intensities of the inputs, confirming that SpliCER is able to learn features of the individual markers more effectively than standard self-supervised methods.

\begin{figure*}[t]
\centering
\begin{subfigure}{0.25\textwidth}
    \centering
    \includegraphics[width=\textwidth]{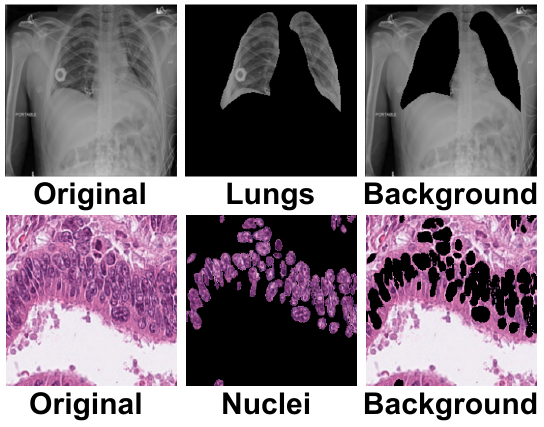}
    \caption{}
    \label{fig:segmentation-image}
\end{subfigure}
\begin{subfigure}{0.6\linewidth}
    \centering
    \scalebox{1}{
    \begin{tabular}{cccc|c}
    \toprule
        \multirow{2}{*}{Loss} & \multirow{2}{*}{Method} & \multicolumn{2}{c}{Histology} & X-ray \\
        & & NCT & Camelyon & PneumoniaMNIST \\
    \midrule
        \multirow{2}{*}{VICReg} & SpliCER & \B 0.9372 & \B 0.8230 & \B 0.8718 \\
        % & $\Sigma$-JE & 0.9209 & 0.7763 \\
        % JE-Split & \cmark & 0.8168 & \\
        % JE & \cmark & 0.8979 & \\
        & Baseline & 0.8855 & 0.6822 & 0.8446 \\
    \midrule
        \multirow{2}{*}{SimCLR} & SpliCER & \B 0.9368 & \B 0.8674 & \B 0.9119 \\
        % & $\Sigma$-JE & 0.9287 & \B 0.9324 \\
        & Baseline & 0.9067 & 0.8346 & 0.8462 \\
    \bottomrule
    \end{tabular}
    }
    \caption{}
    \label{tab:nct-segmentation-results}
\end{subfigure}
\caption{(a) Sample segmentations for ChestMNIST (top) and NCT (bottom) (b) Classification accuracy on NCT, Camelyon and PneumoniaMNIST tasks}
\label{tab:hovernet}
\end{figure*}

\subsection{Hyperspectral Geodata}

We next assess SpliCER's performance in analysing hyperspectral geodata. We pretrain on either the EuroSAT \cite{helber2019eurosat} or MMEarth \cite{nedungadi2024mmearth} datasets, which contains 13-band hyperspectral imaging from the Sentinel-2 satellite program. These bands each correspond to an interval of the electromagnetic spectrum, and each band contains distinct information about a different aspect of the environment, such as atmospheric conditions or soil moisture. We evaluate the quality of the learned representations on the EuroSAT land use classification task, which requires models to use information from all spectral bands to make effective classifications.

\cref{tab:eurosat-results} shows that the land-use classification accuracy is significantly improved from 85\% to 89-90\% by using SpliCER on the 13-band images. In contrast, restricting the model to only use the RGB channels demonstrates no significant different in performance between the methods, and a worse performance overall of 79.67\% for SpliCER and 79.78\% for the baseline VICReg model. This demonstrates that incorporating the additional bands improves classification accuracy, and that ensuring the model learns from all of these additional channels significantly improves performance compared to baseline SSL approaches. We see an even greater increase in performance for SpliCER above baseline when pretraining on MMEarth, with accuracy improved from 50/60\% to 70/71\% for VICReg/SimCLR respectively, indicating that learning more complex features makes models more robust to distribution shift.

\subsection{Segmentation Masks as Image Deconstruction}

Until now the images used for each task have been naturally predisposed to deconstruction due to their compositional design. This is not necessary for SpliCER to provide improved performance, however, as we can use segmentation masks to deconstruct images. In many cases, we have prior knowledge that features of a particular part of the image should be learned, and this can be leveraged with segmentation to improve the quality of the learned representations. By using a segmentation to deconstruct an image into a region of interest and its complement, the model is still able to learn any image features, is no longer able to neglect either segmented component.

\subsubsection{Histopathology Nuclei Segmentation}

It has been shown that histopathology models routinely ignore cell nuclei, instead learning simpler, higher-variance features related to the connective tissue \cite{farndale2023trident,farndale2024synthetic}. Here we use HoVer-Net \cite{graham2019hover} to generate nuclei segmentation masks for each image in the NCT-CRC-100K dataset \cite{kather2019predicting}, which we then use to generate images containing only nuclei, or only connective tissue, as shown in \cref{fig:segmentation-image}. SpliCER can then be used to distil information from both paired images, with the goal of learning features from the nuclei as well as the connective tissue. The simple features here are still useful for downstream prediction, so it is no longer desirable to disregard these. We use synthetically generated segmentation masks to demonstrate that there is no need for exhaustive manual annotation, particularly with the availability of generalist segmentation models such as \emph{Segment Anything} \cite{kirillov2023segment}. There is also a very real biological application of this task, as representation learning is increasingly being used for biological discovery in histopathology \cite{bahadir2024artificial}, but could be hampered by models only learning simple features.

We evaluate on the NCT tissue type classification and Camelyon metastasis detection \cite{bandi2018detection} tasks. Camelyon features a training set of images from three hospitals, and a test set from a different hospital. This results in a distribution shift due to the significant differences in the imaging artefacts between sets. To achieve better performance on Camelyon, models must learn biologically robust features that can generalise beyond the training set. \cref{tab:hovernet} demonstrates that SpliCER achieves a greater accuracy than the baseline on both tasks: 94/94\% on NCT -- a 5 percentage point improvement over baseline, and 82/87\% on Camelyon, compared to a baseline of 68/83\% for VICReg/SimCLR respectively. We evidence in \cref{tab:additional-nct-segmentation-results} that the background features are useful to learn for these tasks, as models paired with only the nuclei mask have worse downstream performance.

\subsubsection{X-Ray Lung Segmentation}

Chest radiographs are typically used to identify very subtle changes in the body, and require extensive training to use effectively. A slight change in brightness can indicate the presence of disease, which may be lost to standard self-supervised models. Radiologists will typically look at the expected site of disease to identify these changes, but the context of the surrounding tissue is also important for identifying disease. We use the HybridGNet model \cite{gaggion2022improving} to generate lung segmentation masks for the ChestMNIST \cite{yang2023medmnist} dataset. These are then used in SpliCER with to direct the model to complex features specifically in the lungs, which are the primary site of the disease. We evaluate the models' performance on PneumoniaMNIST, which is a binary classification task prediction the presence of pneumonia. SpliCER achieves a significantly improved accuracy, with 87/91\% compared to the baseline of 84/85\% for VICReg and SimCLR respectively.
\section{Discussion}

Shortcut solutions are a key issue for self-supervised learning approaches. In the absence of a supervisory signal to enforce the learning of complex features, we have shown that models can default to learning shortcut solutions. We have shown that SpliCER integrates well with existing self-supervised losses, and is highly effective for training models to learn complex features, enabling many more settings to make use of available prior knowledge, particularly multiplex images and those with segmentation masks available. With the development of generalist segmentation models such as \emph{Segment Anything} \cite{kirillov2023segment}, there are a wide range of fields where SpliCER could be used.

SpliCER can be easily adapted to suit different use cases. If a particular channel is known to be more informative, then more features can be allocated to that channel, or if a channel is known to not be informative, then it can be allocated fewer features or omitted altogether. Different losses could be applied for each embedding chunk, allowing customisation of the learned features.

\paragraph{Limitations.}
Regardless of the pretraining method, the classifier head used for downstream tasks is still susceptible to shortcut solutions. If the encoder has simple and complex features but the simple features occur more frequently or are more predictive, these are the features which will be used by the classifier, even if there is value in the complex features. There has been considerable existing work on mitigating simplicity bias \cite{vasudevamitigating,tiwari2023overcoming}. These techniques could be combined with SpliCER, potentially creating a complementary effect in further reducing simplicity bias.

SpliCER is also highly dependent on the quality of the image deconstruction. There may not always be an easy way to segment images, or we may lack the knowledge to do so. Additionally, SpliCER could in theory negatively impact features which span multiple components, as their signal may be impacted by being separated.
\section*{Acknowledgements}

The authors would like to thank Joshua Roche for the insightful discussions about representation learning for radiology, and Baptiste Brauge, Jack McCowan, and Alex Young for their advice on gating strategies.

{
    \small
    \bibliographystyle{ieeenat_fullname}
    \bibliography{main}
}

\clearpage
\setcounter{page}{1}
\maketitlesupplementary

\newcommand{\beginsupplement}{
    \setcounter{section}{0}
    \renewcommand{\thesection}{S\arabic{section}}
    \setcounter{equation}{0}
    \renewcommand{\theequation}{S\arabic{equation}}
    \setcounter{table}{0}
    \renewcommand{\thetable}{S\arabic{table}}
    \setcounter{figure}{0}
    \renewcommand{\thefigure}{S\arabic{figure}}
    \newcounter{SIfig}
    \renewcommand{\theSIfig}{S\arabic{SIfig}}}

\beginsupplement

\begin{figure*}
\begin{subfigure}{0.45\textwidth}
    \centering
    \includegraphics[width=0.9\linewidth]{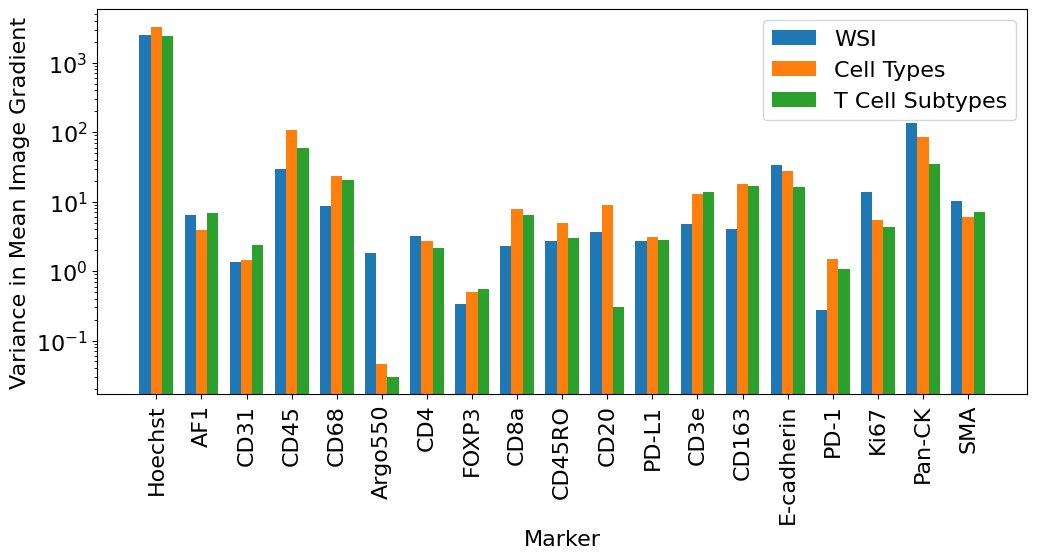}
    \caption{}
    \label{fig:multiplex-variance}
\end{subfigure}
\begin{subfigure}{0.54\textwidth}    
\centering
\scalebox{0.75}{\begin{tabular}{ccccc|ccc}
\toprule
    \multirow{2}{*}{Marker} & \multicolumn{4}{c}{Cell Types} & \multicolumn{3}{c}{T Cells} \\%& \multicolumn{2}{c}{PD-1} \\
    & Tumour & B Cell & Macrophage & T Cell & Treg & CD4$^+$ & CD8$^+$ \\%& CD8$^+$PD-1$^+$ & CD8$^+$PD-1$^-$\\
\midrule
CD45 & - & + & + & + & + & + & + \\% & + & + \\
CD20 & - & + & & - & - & - & - \\% & - & - \\
CD3e & - & - & - & + & + & + & + \\% & + & + \\
CD4 & - & - & & & + & + & - \\% & - & - \\
CD8a & - & - & & & - & - & + \\% & + & + \\
CD45RO & - & & & - & - & - & - \\% & - & - \\
FOXP3 & - & - & & & + & - & \\% & & \\
Pan-CK & + & - & - & - & - & - & - \\% & - & - \\
CD31 & - & - & - & - & - & - & - \\% & - & - \\
CD68 & & & + & & & & \\% & & \\
SMA & - & - & - & - & - & - & - \\% & - & - \\
% PD-1 & & & & & & & & + & - \\
\bottomrule
\end{tabular}
}
\caption{}
\label{tab:gating}
\end{subfigure}

\begin{subfigure}{\textwidth}
\centering
\begin{tabular}{ccccc|ccccc}
\toprule
    \multirow{2}{*}{Task} & \multirow{2}{*}{Method} & \multicolumn{3}{c}{Layers} & \multicolumn{5}{c}{Dimension} \\
    & & 1 & 2 & 3 & 512 & 1024 & 2048 & 4096 & 8192 \\
\midrule
    \multirow{3}{*}{\makecell{Cell Types\\(Simple)}} & SpliCER & \B 0.8840 & 0.8845 & \B 0.9003 & \B 0.8683 & 0.8720 & 0.8730 & \B 0.8898 & \B 0.9003 \\
    & $\Sigma$-JE & 0.4315 & 0.5113 & 0.6715 & 0.4697 & 0.5788 & 0.5785 & 0.6628 & 0.6715 \\
    & Baseline & 0.8687 & \B 0.8850 & 0.8820 & 0.8625 & \B 0.8935 & \B 0.8755 & 0.8832 & 0.8820 \\
\midrule
    \multirow{3}{*}{\makecell{T Cells\\(Complex)}} & SpliCER & \B 0.7589 & \B 0.7933 & \B 0.8163 & \B 0.7779 & \B 0.7889 & \B 0.7996 & \B 0.8166 & \B 0.8163 \\
    & $\Sigma$-JE & 0.4158 & 0.4545 & 0.5222 & 0.3958 & 0.4471 & 0.4522 & 0.5152 & 0.5222 \\
    & Baseline & 0.5929 & 0.5882 & 0.5665 & 0.5709 & 0.5115 & 0.5939 & 0.6185 & 0.5665 \\
    \bottomrule
\end{tabular}
\caption{}
\label{tab:projector-ablations}
\end{subfigure}
\caption{(a) Average intra-image variance of channels in Orion-CRC subsets (b) Gating strategy used to create tasks for downstream analysis (c) Projector ablations on Orion-CRC}
\label{}
\end{figure*}

\section{Extended Dataset Descriptions}
\label{sec:datasets}

\subsection{MNIST-CIFAR}

The MNIST-CIFAR dataset is the standard test used to assess simplicity bias, and is designed to assess models' ability to learn complex features \cite{shah2020pitfalls,morwani2023simplicity,tiwari2023overcoming}. The dataset consists of digits from MNIST \cite{lecun1998gradient} concatenated to images from the CIFAR10 dataset \cite{krizhevsky2009learning}, as shown in \ref{fig:mnistcifar-examples}. The MNIST images contain very simple features, and models typically achieve close to 100\% accuracy on this test. In contrast, the features in CIFAR10 are more complex, and are not as easily learned. After training on the concatenated images, the level of simplicity bias can be assessed by randomising either the MNIST (denoted rMNIST) or CIFAR (denoted rCIFAR) component. The model's difference in performance between rNONE (no randomisation) and rMNIST determines the level of simplicity bias. If only the simple MNIST features have been learned, the performance will drop from near perfect to random guessing, while better accuracy indicates complex features have been learned from the CIFAR images. Conversely, if there is a performance drop on rCIFAR, the model has neglected some MNIST features and only learned the complex features.

Even if the encoder has learned both simple and complex features, the MNIST features will inevitably be more predictive and have greater variance. This means that the classifier will be biased towards the simpler MNIST features. A better test is therefore to train the classifier on the rMNIST dataset. This means that the classifier can only use the complex CIFAR features for classification, and directly assesses the complexity of the learned representations. In this setting we begin to see differences depending on the pretraining method. 

\subsection{EuroSAT}

EuroSAT \cite{helber2019eurosat} is a well-established benchmark in geospatial image analysis. It consists of 27,000 $64\times64$px images at a resolution of 10 meters per pixel from the Sentinel-2 satellite, which is a part of the European Space Agency's Copernicus program. The images contain 13 spectral bands, which each correspond to a different region of the electromagnetic spectrum. These bands each contain different information about the land use and environmental properties of the region depicted in the image, such as cloud cover, soil moisture, and biomass estimation. The downstream task associated to EuroSAT is 10 class classification of the categories \emph{Annual Crop}, \emph{Forest}, \emph{Herbaceous Vegetation}, \emph{Highway}, \emph{Industrial}, \emph{Pasture}, \emph{Permanent Crop}, \emph{Residential}, \emph{River}, \emph{Sea Lake}.

\subsection{MMEarth}

The MMEarth dataset \cite{nedungadi2024mmearth} is a large multimodal dataset containing data from 12 modalities for 1.2 million locations. We only utilise the Sentinel-2 data, which consists of 1.2 million $64\times64$px patches, with the same 13 spectral bands as EuroSAT. As there is no associated downstream task with MMEarth, we use EuroSAT to evaluate models trained on MMEarth.

\subsection{NCT-CRC-100K}

The NCT-CRC-100K dataset \cite{kather2019predicting} is a widely used benchmark dataset in computational pathology. It features 100,000 $224\times224$px patches taken from 86 patients with colorectal cancer, approximately evenly split into 9 categories:  \emph{Adipose}, \emph{Background}, \emph{Debris}, \emph{Lymphocytes}, \emph{Mucus}, \emph{Smooth Muscle}, \emph{Normal Colon Mucosa}, \emph{Cancer-Associated Stroma}, and \emph{Colorectal Adenocarcinoma Epithelium}. There is also an associated test dataset containing 7180 patches from a separate 25 patients.

\subsection{Camelyon}

To assess the robustness of the models trained on NCT-CRC-100K to distribution shift, we use the WILDS Camelyon17 dataset \cite{wilds2021}. This is a variant of the original Camelyon17 \cite{bandi2018detection} dataset, which features 1400 whole slide images of H\&E stained lymph node sections. The dataset is constructed to assess robustness to distribution shift, with slides from three hospitals used for training, and slides from a different hospital used for testing. The dataset features $96\times96$px patches with a train/test split of 179,394/146,722. We resize the patches to $224\times224$px for consistency with NCT-CRC-100K. The evaluation task associated with Camelyon is binary classification of the presence of metastasis.

\subsection{Orion-CRC}

To the best of our knowledge, there are no existing benchmarks for machine learning on multiplex images. We therefore construct two datasets by subsampling the publicly available Orion-CRC dataset \cite{lin2023high}. Orion-CRC contains 41 pathology slides, images with both H\&E staining and 17-plex multiplex immunofluorescence. The markers used were \emph{Hoechst}, \emph{CD31}, \emph{CD45}, \emph{CD68}, \emph{CD4}, \emph{FOXP3}, \emph{CD8$\alpha$}, \emph{CD45RO}, \emph{CD20}, \emph{PD-L1}, \emph{CD3e}, \emph{CD163}, \emph{E-cadherin}, \emph{PD-1}, \emph{Ki67}, \emph{Pan-CK}, and \emph{SMA}, and the images contain an autofluorescence channel (AF1) and a negative control (Argo550).

Using the segmentation mask provided for the CRC02 slide, we randomly subsampled 1\% of all cells to create a dataset of 12,606 cells, which were extracted in $64\times64$px patches. We then created two zero-shot downstream evaluation tasks from CRC01, following the same protocol but randomly selecting 2000 cells from each desired cell type, split evenly into train and test sets. Cells were defined by the gating strategy in \cref{tab:gating} For the \emph{Cell Types} task, this was \emph{Tumour}, \emph{B Cells}, \emph{Macrophages} and \emph{T Cells}. This task requires only simple features, as cells have very different marker profiles and are mostly morphologically distinct (B and T cells are less morphologically distinct than the others but express different markers). For the \emph{T Cells} task, we randomly selected \emph{CD4$^+$}, \emph{CD8$^+$}, and \emph{regulatory T cells (Tregs)} as our cell types, which are all T cell subsets. This requires more complex features to be learned, as these channels have lower variance and are morphologically indistinguishable. 

To analyse which channels models are most likely to focus on, we analyse the variance in the image gradients of each channel. Hoechst has the largest variance by a significant margin, whereas some stains, such as FOXP3, can have three orders of magnitude less variance. This means that models are likely to ignore features in FOXP3, which are uncommon and therefore can incur large penalties in the pretraining loss. In contrast, features in Hoechst will be readily learned. We also observe large differences between intensities of different subsets of cells. For example, there is much greater variance among T cells than among all cells in the sample, as FOXP3 is a marker for Tregs. In contrast, there is considerably less variance in CD20 and pan-CK, as T cells do not express the these proteins, so any expression is likely to be either off-target staining or a T cell being adjacent to a tumour cell or B cell.

\subsection{MedMNIST}

Inspired by the MNIST handwritten digits dataset \cite{lecun1998gradient}, MedMNIST \cite{yang2023medmnist} is a collection of 18 standardised biomedical datasets which have been preprocessed to ensure ease of use. The images are available in 28, 64, 128, and 224px sizes, and each dataset has one or more associated classification tasks. In this work we use both ChestMNIST and PneumoniaMNIST, which feature chest radiographs from the NIH-ChestXray14 dataset \cite{wang2017chestx} and a dataset from Kermany \etal \cite{kermany2018identifying}.

ChestMNIST features 112,120 frontal-view X-rays from 30,805 patients, split into train, validation, and test sets of size 78468, 11,219, ad 22,433 respectively. PneumoniaMNIST features 5856 frontal-view chest X-rays, with train/validation/test splits of 4708/524/624 respectively.

\begin{figure*}
\centering
\begin{subfigure}{0.5\textwidth}
    \centering
    \begin{tabular}{cccc}
    \toprule
        Method & \makecell{Cell Types\\(Simple)} & \makecell{T Cells\\(Complex)} \\
    \midrule
        SpliCER & 0.9003 & \B 0.8163 \\
        Multiple Encoders & 0.8928 & 0.8029 \\
        Double Hyperstack Branch & \B 0.9035 & 0.8116 \\
        Baseline + chunking & 0.8827 & 0.6069 \\
        % Channel Weighting & \\
    \bottomrule
    \end{tabular}
    \caption{}
    \label{tab:alternative-ablations}
\end{subfigure}
\hspace{0.5cm}
\begin{subfigure}{0.45\textwidth}
    \centering
    \begin{tabular}{ccc}
    \toprule
        Method & \makecell{Cell Types\\(Simple)} & \makecell{T Cells\\(Complex)} \\
    \midrule
        Baseline & 0.8820 & 0.5665 \\
        + additional branches & 0.6578 & 0.5545 \\
        + distinct projectors & 0.6715 & 0.5222 \\
        + chunking (SpliCER) & \B 0.9003 & \B 0.8163 \\
    \bottomrule
    \end{tabular}
    \caption{}
    \label{tab:constructive-ablations}
\end{subfigure}
\caption{(a) Ablations on Orion-CRC tasks showing possible adaptations to SpliCER (b) Ablations on Orion-CRC tasks breakig down each element of SpliCER}
\label{fig:ablations}
\end{figure*}

\section{Training and Implementation Details}
\label{sec:training}

Models were trained for 100 epochs with a batch size of 256. The exception to this are the MMEarth models, which were trained for 100 epochs. The backbone encoder was a ResNet-18 \cite{he2016deep} for all tasks except the histology segmentation training, where ResNet-50 was used. Primary models were always trained from random initialisation and were appended with three layer projection heads with layer size 8192, batch normalisation between layers, ReLU activations, and a linear final layer. \cref{tab:projector-ablations} shows the results are robust to this choice. For SpliCER, the primary projection head was split into evenly distributed chunks, and the projection head was scaled accordingly by the number of channels. For example, there are 19 channels in the Orion-CRC dataset, so each projection head had output dimension 431. In the Orion-CRC and MMEarth examples, a single paired encoder was used for all single-channel inputs -- two in total. For the MNIST-CIFAR and histology segmentation examples, a separate encoder was used for each branch, giving three encoders in total. An Adam optimiser \cite{kingma2014adam} with a warmup cosine learning rate was used, warming up the learning rate from 0 to $10^{-4}$ over the first 10\% of epochs. There were no differences in the hyperparameters used for SpliCER and any baselines.

Two loss functions were tested: VICReg \cite{bardesvicreg} and SimCLR \cite{chen2020simple}. VICReg is a non-contrastive loss which contains terms to regularise the variance and covariance of each embedding, and enforcing pairwise invariance between embeddings. SimCLR is a contrastive method using the InfoNCE loss \cite{oord2018representation}, which is also used in methods such as CLIP \cite{radford2021learning}. For VICReg, we used the standard parameters $\lambda=\mu=25,\nu=1$, and for SimCLR we used the temperature $\tau=0.5$.

For downstream tasks, encoder weights were frozen and a linear classifier with softmax activation was used to perform classification. Classifiers were trained for 100 epochs with data augmented with the same regime used for pretraining. All metrics reported are mean per-class accuracy.

\begin{table}[]
\centering
\caption{Effect of Normalisation}
\label{tab:orion-normalisation}
\scalebox{0.9}{
\begin{tabular}{cccccc}
    \toprule
    Task & Method & None & \makecell{Normalise-\\only} & \makecell{Clip \&\\Normalise} \\
    \midrule
    \multirow{3}{*}{\makecell{Cell Types\\(Simple)}} & SpliCER & \B 0.9003 & \B 0.7550 & 0.6710 \\
    & $\Sigma$-JE & 0.6715 & 0.6242 & 0.6018 \\
    & Baseline & 0.8820 & 0.5940 & \B 0.7285 \\
    \midrule
    \multirow{3}{*}{\makecell{T Cells\\(Complex)}} & SpliCER & \B 0.8163 & 0.3558 & \B 0.6789 \\
    & $\Sigma$-JE & 0.5222 & 0.3525 & 0.4812 \\
    & Baseline & 0.5665 & \B 0.3598 & 0.5582 \\
    \bottomrule
\end{tabular}
}
\end{table}

\section{Ablations}
\label{sec:ablations}

\subsection{Normalisation and Clipping}

We assess the effect of normalisation to demonstrate that simply normalising the inputs does not address feature variance. We employ two different normalisation schemes: normalisation, and normalisation with clipping. Multiplexed immunofluorescence typically have a long tail and background noise, so a standard approach is to clip the bottom and top 5\% of values in each channel before normalising to zero mean and unit variance \cite{wolflein2023whole}.

We find that the improved performance of SpliCER cannot be explained by implicit normalisation. In \cref{tab:orion-normalisation}, we demonstrate that normalisation of the data only reduces performance, and SpliCER on unnormalised data consistently outperforms the other methods. We postulate than normalisation is harmful due to condensing the majority of values to a very small interval, as these images are dominated by a small number of large values.

\begin{table*}[]
    \centering
    \caption{MNIST-CIFAR results for original classes (0: car, 1: truck) and for new, out-of-distribution classes (0: bird, 1: cat)}
    \label{tab:mnist-cifar-shared-latent-space}
    \begin{tabular}{ccccccc}
        \toprule
        \multirow{2}{*}{Method} & \multicolumn{3}{c}{Original Classes} & \multicolumn{3}{c}{New Classes} \\
        & rNONE & rMNIST & rCIFAR & rNONE & rMNIST & rCIFAR \\
        \midrule
        SpliCER & 0.9980 & 0.6840 & 0.9974 & 0.9974 & 0.6638 & 0.9980 \\
        $\Sigma$-JE & 0.9990 & 0.7485 & 0.9990 & 0.9990 & 0.5855 & 0.9985 \\
        Baseline & 0.9969 & 0.6275 & 0.9964 & 0.9959 & 0.6265 & 0.9964 \\
        \bottomrule
    \end{tabular}
\end{table*}

\begin{table}
    \centering
    \caption{Evidence of usefulness of background features from the histology segmentation tasks}
    \label{tab:additional-nct-segmentation-results}
    \scalebox{0.9}{
    \begin{tabular}{cccc}
    \toprule
        Loss & Method & NCT & Camelyon \\
    \midrule
        \multirow{4}{*}{VICReg} & SpliCER & \B 0.9372 & \B 0.8230 \\
        & TriDeNT (Nuclei Paired) & 0.9209 & 0.7763 \\
        % JE-Split & \cmark & 0.8168 & \\
        & Baseline (Nuclei Paired) & 0.8979 & 0.8013 \\
        & Baseline & 0.8855 & 0.6822 \\
    \bottomrule
    \end{tabular}
    }
\end{table}

\subsection{Alternative Approaches}

In \cref{fig:ablations} we investigate alternative design choices which could be made in SpliCER. We observe marginally worse performance using a separate encoder for each branch and using one encoder for all branches (in both cases there is a separate encoder for the primary branch). This could be due to a single shared encoder essentially seeing 19$\times$ as many training examples. Intuitively, there could be scenarios where it was more important to use different encoders, such as when we have prior knowledge that different branches contain very different input images, or if there are different input dimensions.

We also experiment with using a second hyperstack, following TriDeNT \cite{farndale2023trident}. This does not appear to significantly affect the performance. This is likely because all of the information in the hyperstack is also present in at least one branch of the paired data, so there are no features which could be neglected by not being paired. We find that the embedding chunking mechanism alone is insufficient to achieve the performance gains of SpliCER. If there are not multiple branches, the chunking is redundant, and is essentially equivalent to halving the size of the projector, as the remaining half only receives signal from the variance and covariance loss terms.

In \cref{tab:constructive-ablations}, we break down SpliCER into its constituent parts to assess the impact of each. We see that na\"ively adding branches makes the downstream performance considerably worse. This is not significantly affected by adding distinct projectors, but the addition of chunking to create SpliCER achieves significant performance gains on the complex T Cells task, and restores performance on the simple Cell Types task.

Finally, in \cref{tab:additional-nct-segmentation-results}, we confirm that the background provides useful information in the NCT and Camelyon tasks. We train both a baseline VICReg model and a TriDeNT model with just the nuclei as paired data. This improves performance compared to the unpaired baseline, but fails to reach the performance of SpliCER. This indicates that by incorporating features from the background SpliCER has a better set of learned features for these downstream tasks.

% Construct a mask overlap task (do cd3/cd8 occur in same region of cell etc) to assess whether single channel model compares to splicer

\section{Additional Analysis}

\subsection{TriDeNT Optimisation Problem}

TriDeNT \cite{farndale2023trident} has been proposed as a method to address this issue, balancing learning primary and privileged features. This allows the model to use the privileged information only so far as it is useful, and neglect it if not. Concretely, the optimisation problem becomes
\begin{equation}
\label{eq:trident-lagrange}
\begin{split}
    \mathcal{L} & = \sum_{\substack{i, j \in \{1,2,*\}\\i\neq j}}\mathcal{L}_{i\rightarrow j}(\lambda_{ij}) \\
    & = \sum_{\substack{i, j \in \{1,2,*\}\\i\neq j}}I(x_i;e_i|x_j)-\lambda_{ij} I(x_j;e_i).
\end{split}
\end{equation}
The increased degrees of freedom allow $\lambda_{ij}$ to be implicitly adjusted to cater to the privileged information, such that information which may be considered superfluous in the unprivileged formulation can now be learned if it is shared between the primary and privileged data. This mitigates the effects of strong augmentation on the features learned, as features which would otherwise be lost to augmentation can now have a strong supervisory signal from the privileged information. The loss function for TriDeNT does not achieve this optimally, as the optimisation process still penalises superfluous information between primary views, even if the same information is not superfluous for the primary/privileged terms, and vice-versa.

\subsection{Mapping Inputs to One Shared Latent Space Overconstrains Models}

We note that there is a significant difference in the performance of $\Sigma$-JE between the MNIST-CIFAR and the other tasks. This can be understood in the context of the multi-view assumption (\cref{eq:multi-view-assumption}). There is no restriction on models' ability to learn any features shared between all branches, as these are present in all views. Features of the primary input which are not shared between all branches inherently must be shared with at least one branch. When there are two privileged branches, as in MNIST-CIFAR, the supervisory signal for each feature is half coming from the branch with the feature, and half coming from the branch without the feature. Therefore, the supervisory signal for that feature to the main encoder can still be reasonably strong. In contrast, if there are a large number of branches without the feature (or a strongly correlated feature), the signal is considerably diluted by unrelated features or collapsed dimensions, making the feature difficult to learn unless its underlying signal is very strong. This is discussed further in \cref{sec:shared-latent-space-supp}

\begin{table*}
    \centering
    \caption{Full MNIST-CIFAR Results}
    \label{tab:full-mnist-cifar}
    \scalebox{0.8}{\begin{tabular}{cccccccccccc}
    \toprule
         & & & \multicolumn{3}{c}{rNONE (All)} & \multicolumn{3}{c}{rMNIST (Complex)} & \multicolumn{3}{c}{rCIFAR (Simple)} \\
         Loss & Paired Image & Method & rNONE & rMNIST & rCIFAR & rNONE & rMNIST & rCIFAR & rNONE & rMNIST & rCIFAR \\
    \midrule
           \multirow{10}{*}{VICReg} & \multirow{2}{*}{NONE} & TriDeNT &   0.9954 &  0.5165 &  0.9939 &   0.6765 &  0.6230 &  0.5434 &   0.9929 &  0.5160 &  0.9939 \\
           &      & Baseline &   0.9908 &  0.5160 &  0.9908 &   0.6061 &  0.6410 &  0.4673 &   0.9918 &  0.5150 &  0.9918 \\
           \cmidrule{2-12}
           & \multirow{3}{*}{MNIST} & SpliCER &   0.9974 &  0.5150 &  0.9985 &   0.6459 &  0.6425 &  0.5066 &   0.9974 &  0.5150 &  0.9974 \\
           & & TriDeNT &   0.9985 &  0.5160 &  0.9985 &   0.3668 &  0.4990 &  0.3449 &   0.9985 &  0.5160 &  0.9980 \\
           &      & Baseline &   0.9980 &  0.5160 &  0.9980 &   0.6245 &  0.5270 &  0.6071 &   0.9985 &  0.5165 &  0.9980 \\
           \cmidrule{2-12}
           & \multirow{3}{*}{CIFAR} & SpliCER &   0.9888 &  0.5145 &  0.9878 &   0.7199 &  0.6840 &  0.5235 &   0.9867 &  0.5150 &  0.9852 \\
           & & TriDeNT &   0.8699 &  0.6155 &  0.7577 &   0.6776 &  0.6645 &  0.5010 &   0.8291 &  0.5400 &  0.8041 \\
           &      & Baseline &   0.7372 &  0.6745 &  0.5847 &   0.6883 &  0.6830 &  0.4827 &   0.6719 &  0.5020 &  0.7020 \\
           \cmidrule{2-12}
           & \multirow{2}{*}{BOTH} & SpliCER &   0.9964 &  0.5155 &  0.9964 &   0.6781 &  0.6675 &  0.4893 &   0.9969 &  0.5170 &  0.9969 \\
           & & $\Sigma$-JE & 0.9934 &  0.4855 & 0.9929 & 0.6071 & 0.6530 & 0.4510 & 0.9934 & 0.4855 & 0.9934 \\
           \cmidrule{1-12}
           \multirow{10}{*}{SimCLR} & \multirow{2}{*}{NONE} & TriDeNT &   0.9974 &  0.5165 &  0.9974 &   0.6515 &  0.6390 &  0.5158 &   0.9974 &  0.5165 &  0.9969 \\
           &      & Baseline &   0.9974 &  0.5160 &  0.9964 &   0.6301 &  0.6260 &  0.5179 &   0.9974 &  0.5170 &  0.9964 \\
           \cmidrule{2-12}
           & \multirow{3}{*}{MNIST} & SpliCER &   0.9959 &  0.5175 &  0.9944 &   0.6224 &  0.6100 &  0.5036 &   0.9964 &  0.5170 &  0.9949 \\
           & & TriDeNT &   0.9980 &  0.5155 &  0.9985 &   0.5776 &  0.5670 &  0.4995 &   0.9985 &  0.5160 &  0.9985 \\
           &      & Baseline &   0.9980 &  0.5155 &  0.9974 &   0.5643 &  0.5875 &  0.4852 &   0.9980 &  0.5155 &  0.9974 \\
           \cmidrule{2-12}
           & \multirow{3}{*}{CIFAR} & SpliCER &   0.9908 &  0.5140 &  0.9893 &   0.7128 &  0.6940 &  0.5117 &   0.9903 &  0.5130 &  0.9888 \\
           & & TriDeNT &   0.9408 &  0.5330 &  0.9250 &   0.7097 &  0.6760 &  0.5255 &   0.9066 &  0.5020 &  0.9219 \\
           &      & Baseline &   0.8224 &  0.5950 &  0.7362 &   0.7005 &  0.6880 &  0.5158 &   0.7566 &  0.5030 &  0.7561 \\
           \cmidrule{2-12}
           & \multirow{2}{*}{BOTH} & SpliCER &   0.9964 &  0.5165 &  0.9949 &   0.6469 &  0.6735 &  0.4781 &   0.9969 &  0.5165 &  0.9964 \\
           & & $\Sigma$-JE & 0.9959 & 0.4840 & 0.9964 & 0.6592 & 0.6840 & 0.4883 & 0.9969 & 0.4845 & 0.9959 \\
           \bottomrule
\end{tabular}
}
\end{table*}

\subsection{Use-Case for Mapping All Channels into a Shared Latent Space}
\label{sec:shared-latent-space-supp}

There is a highly specific use-case for mapping all channels into a shared latent space: when a downstream task is known and all channels are strongly associated with the downstream label, there is some utility in training aligning all embeddings. As there is no requirement that the features learned by each branch are necessarily the same, there may be features that are highly correlated but bare no resemblance to each other, such as associating the shape of 0s and properties of cars. These can be mapped to the same embedding element, but provide no meaningful information about the other branch. In the extreme case, where the branches are not related to each other at all (such as MNIST-CIFAR), this causes collapse to any shared feature. For MNIST-CIFAR, the shared feature is the correspondence between e.g. 0 and car, or 1 and truck. In this case, we find that SSL training with unrandomised MNIST and CIFAR in a shared latent space causes collapse to a low-rank solution with very high correlation to the label. This essentially reduces the problem to supervised learning with feature regularisation, as the signal from the additional branch serves only to differentiate cars from trucks, or 0s from 1s, with no information about the features learned.

This approach could be helpful in very specific cases, however, it is not useful for general representation learning, as the resultant representations are not robust to domain shift, much like supervised learning. We show in \cref{tab:mnist-cifar-shared-latent-space} that changing the image classes used as CIFAR inputs considerably reduces the performance of the $\Sigma$-JE model, while SpliCER can generalise well. This is because SpliCER creates a supervisory signal that encourages the learning of generic image features, which $\Sigma$-JE lacks. 

\cref{tab:full-mnist-cifar} shows the results for all models evaluated on all datasets, including those they were not trained on. We see that this reproduces the results of \cite{shah2020pitfalls}, where models without paired data collapse to random accuracy on rMNIST and achieve good results on rNONE and rCIFAR. Interestingly, we observe that models paired with CIFAR can perform on rMNIST even when trained on rNONE. This implies that they have neglected simple MNIST features in favour of learning CIFAR features, as MNIST features would likely be used by the classifier head when training on rNONE. We propose that SpliCER always performs poorly in this setting because it has learned both MNIST and CIFAR features.

\end{document}